\documentclass[12pt]{article}
\usepackage{amsmath}
\usepackage{amsfonts}
\usepackage{amssymb}
\usepackage{graphicx}
\usepackage{hyperref}
\usepackage{geometry}
\usepackage{booktabs}
\usepackage{multirow}
\usepackage{indentfirst}
\usepackage{parskip}
\usepackage{fullpage}
\usepackage{algpseudocode}
\usepackage{caption} 
\usepackage{cleveref}

\hypersetup{
    colorlinks=true,
    linkcolor=blue,
    filecolor=magenta,      
    urlcolor=blue,
    citecolor=blue
}

\title{Gradient Descent Efficiency Index}
\author{Aviral Dhingra\\
\small\href{mailto:aviral.dhingra.2008@gmail.com}{aviral.dhingra.2008@gmail.com}}
\date{}

\begin{document}

\maketitle

\begin{abstract}
\noindent
Gradient descent is a widely used iterative algorithm for finding local minima in multivariate functions. However, the final iterations often either overshoot the minima or make minimal progress, making it challenging to determine an optimal stopping point. This study introduces a new efficiency metric, \( E_k \), designed to quantify the effectiveness of each iteration. The proposed metric accounts for both the relative change in error and the stability of the loss function across iterations. This measure is particularly valuable in resource-constrained environments, where costs are closely tied to training time. Experimental validation across multiple datasets and models demonstrates that \( E_k \) provides valuable insights into the convergence behavior of gradient descent, complementing traditional performance metrics. The index has the potential to guide more informed decisions in the selection and tuning of optimization algorithms in machine learning applications and be used to compare the ``effectiveness" of models relative to each other.
\end{abstract}

\section{Introduction}

In the field of machine learning, optimizing the training process of models is crucial for achieving high performance while minimizing computational resources. Gradient descent \cite{GradientDescent} is a widely used optimization algorithm due to its simplicity and effectiveness in finding local minima of differentiable functions. However, the efficiency of gradient descent can diminish with large datasets and prolonged training periods, where additional iterations provide negligible improvements. This raises the need for a robust mechanism to identify the optimal stopping point, ensuring efficient use of computational resources.

\begin{figure}[htbp]
\centering
\includegraphics[width=\linewidth]{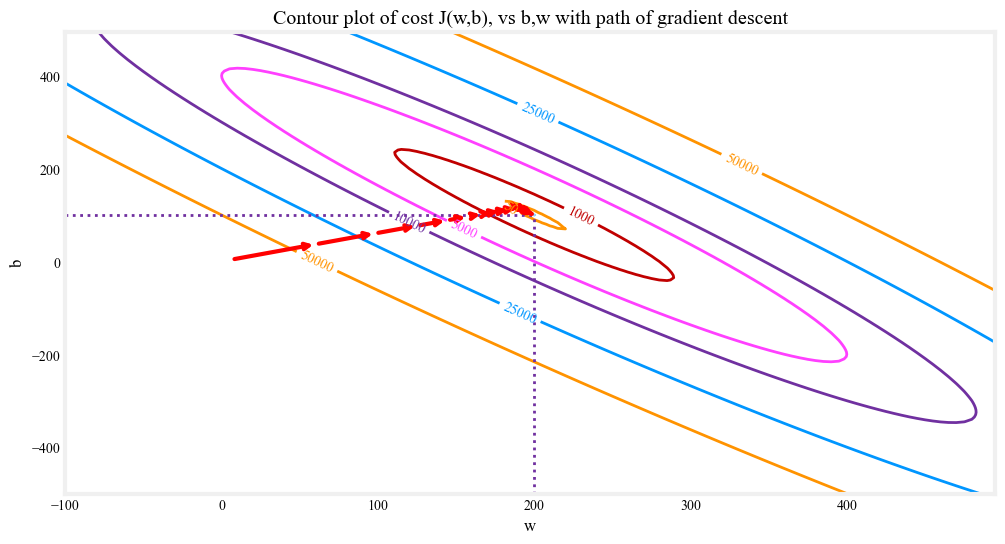}
\caption{Contour plot of cost b,w with path of gradient descent \(J(w,b)\).}
\label{fig:contourplot}
\end{figure}

\hyperref[fig:contourplot]{Fig.~\ref*{fig:contourplot}} is a contour plot that illustrates how each step taken in gradient descent towards a local minimum is smaller than the previous one. This approach quickly returns diminishing results, making the last few steps cost more computationally than they yield in accuracy.

\begin{figure}[htbp]
\centering
\includegraphics[width=\linewidth]{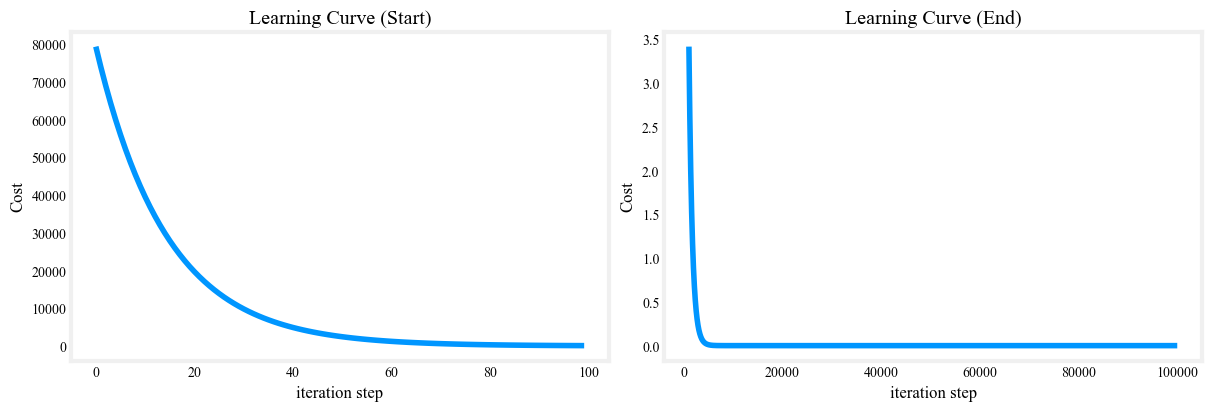}
\caption{Comparison of cost with different domain restrictions.}
\label{fig:learningcurve}
\end{figure}

In \hyperref[fig:learningcurve]{Fig.~\ref*{fig:learningcurve}}, notice how the first 100 steps exponentially decrease (or logarithmically increase) the cost. However, if you zoom out to 100,000 steps, the curve effectively flattens out before 10,000 steps in this particular example.

The ``Gradient Descent Efficiency Index" is a novel ratio between training parameters that includes the relative change in gradient norm, the initial learning rate, the learning decay rate, absolute change in coefficients, and the number of iterations.

Note that throughout this paper, I will use the name ``Gradient Descent Efficiency Index", the short form ``GDEI", and the function \textit{\( E_k \)} interchangeably.

\section{Related Work}

\subsection{Momentum}

Momentum \cite{Momentum} helps accelerate gradient vectors in the right directions, thus leading to faster convergence.

\[
    v_t = \beta v_{t-1} + (1 - \beta) g_t
\]
\[
    \theta_{t+1} = \theta_t - \alpha v_t
\]

where:
\begin{itemize}
    \item \( v_t \) is the velocity vector.
    \item \( \beta \) is the momentum hyperparameter.
    \item \( \alpha \) is the learning rate.
    \item \( g_t \) is the gradient at time step \( t \).
\end{itemize}

\subsection{AdaGrad}

AdaGrad \cite{Adagrad} adapts the learning rate for each parameter based on the past gradients.

\[
    G_t = \sum_{\tau=1}^t g_{\tau}^2
\]
\[
    \theta_{t+1} = \theta_t - \frac{\alpha}{\sqrt{G_t + \epsilon}} g_t
\]

where:
\begin{itemize}
    \item \( G_t \) is the sum of the squares of the past gradients.
    \item \( \alpha \) is the global learning rate.
    \item \( \epsilon \) is a small constant to prevent division by zero.
    \item \( g_t \) is the gradient at time step \( t \).
\end{itemize}

\subsection{RMSProp}

RMSProp \cite{RMSprop} (Root Mean Square Propagation) adjusts the learning rate for each parameter.

\[
    E[g^2]_t = \beta E[g^2]_{t-1} + (1 - \beta) g_t^2
\]
\[
    \theta_{t+1} = \theta_t - \frac{\alpha}{\sqrt{E[g^2]_t + \epsilon}} g_t
\]

where:
\begin{itemize}
    \item \( E[g^2]_t \) is the exponentially decaying average of past squared gradients.
    \item \( \beta \) is the decay rate.
    \item \( \alpha \) is the learning rate.
    \item \( \epsilon \) is a small constant to prevent division by zero.
    \item \( g_t \) is the gradient at time step \( t \).
\end{itemize}

\subsection{Adam}

Adam (Adap) \cite{Adam} combines the advantages of both AdaGrad and RMSProp.

\[
    m_t = \beta_1 m_{t-1} + (1 - \beta_1) g_t
\]
\[
    v_t = \beta_2 v_{t-1} + (1 - \beta_2) g_t^2
\]
\[
    \hat{m}_t = \frac{m_t}{1 - \beta_1^t}
\]
\[
    \hat{v}_t = \frac{v_t}{1 - \beta_2^t}
\]
\[
    \theta_{t+1} = \theta_t - \frac{\alpha \hat{m}_t}{\sqrt{\hat{v}_t} + \epsilon}
\]

where:
\begin{itemize}
    \item \( m_t \) and \( v_t \) are the first and second moment estimates, respectively.
    \item \( \beta_1 \) and \( \beta_2 \) are hyperparameters for the decay rates.
    \item \( \alpha \) is the learning rate.
    \item \( \epsilon \) is a small constant to prevent division by zero.
    \item \( g_t \) is the gradient at time step \( t \).
\end{itemize}

\subsection{Nesterov Accelerated Gradient (NAG)}

Nesterov Accelerated Gradient (NAG) \cite{nesterov1983method} is a variation of the momentum method that anticipates the future position of the parameters. Unlike standard momentum, which calculates the gradient at the current position, NAG first makes a big jump in the direction of the accumulated gradient and then measures the gradient. The update rules are:

\[
v_{k+1} = \gamma v_k + \eta \nabla L(\theta_k - \gamma v_k)
\]
\[
\theta_{k+1} = \theta_k - v_{k+1}
\]

NAG often results in faster convergence compared to standard momentum, especially in settings with high curvature.

\subsection{Other Algorithms and Variants}

In addition to the widely used methods mentioned above, numerous other algorithms and variants have been proposed to address specific challenges in gradient-based optimization\cite{GDLitReview}. These include:

\begin{itemize}
    \item \textbf{SGD with Warm Restarts:} A variant of stochastic gradient descent (SGD) that periodically restarts with a large learning rate to escape local minima.
    \item \textbf{AdaMax:} An extension of Adam that uses the infinity norm instead of the second moment, making it more robust in certain applications.
    \item \textbf{AMSGrad:} A modification of Adam that aims to improve its convergence properties by fixing a flaw in the original algorithm.
    \item \textbf{Nadam:} Combines the Nesterov Accelerated Gradient with Adam, offering a blend of adaptive learning rates and look-ahead gradient calculation.
\end{itemize}

Each of these algorithms contributes uniquely to the field of optimization, providing various trade-offs in terms of convergence speed, stability, and computational cost. However, a common challenge across all these methods is the lack of a standardized metric to measure the efficiency of each iteration. The efficiency score \( E_k \) proposed in this paper aims to address this gap, offering a more detailed perspective on the performance of gradient descent.

\section{Derivation of GDEI}

\subsection{External Parameters}

\subsubsection{Mean Squared Error (MSE)}

The error metric used is the Mean Squared Error (MSE) \cite{MSE}. It is defined as:

\[
E = \frac{1}{n} \sum_{i=1}^{n} (y_i - \hat{y}_i)^2
\]

where:
\begin{itemize}
    \item \( n \) is the number of data points.
    \item \( y_i \) is the actual value of the \( i \)-th data point.
    \item \( \hat{y}_i \) is the predicted value of the \( i \)-th data point.
\end{itemize}

This formula calculates the average of the squared differences between the actual and predicted values, providing a measure of the quality of the model's predictions.

\subsubsection{Proportion of Initial Loss Reduced (\( P_k \))}

The parameter \( P_k \) represents the proportion of the initial error that has been reduced by the \( k \)-th iteration. It is defined as:

\[
P_k = \frac{L_{\text{initial}} - L_k}{L_{\text{initial}}}
\]

where:
\begin{itemize}
    \item \( L_{\text{initial}} \) is the initial mean squared error at the beginning of the optimization process.
    \item \( L_k \) is the current mean squared error at the \( k \)-th iteration.
\end{itemize}

This term quantifies the effectiveness of each iteration in reducing the error, with a higher \( P_k \) indicating greater progress towards minimizing the loss function.

\subsubsection{Absolute Change in Loss Function (\( \Delta_k \))}

The parameter \( \Delta_k \) denotes the absolute change in the loss function, which is the MSE, between consecutive iterations. It is defined as:

\[
\Delta_k = |L_{k-1} - L_k|
\]

where:
\begin{itemize}
    \item \( L_{k-1} \) is the mean squared error at the \((k-1)\)-th iteration.
    \item \( L_k \) is the mean squared error at the \( k \)-th iteration.
\end{itemize}

This term captures the stability of the optimization process. Large values of \( \Delta_k \) suggest instability, which can indicate inefficient steps or overly aggressive learning rates.

\subsection{Formula \& Theoretical Explanation}

The efficiency score \( E_k \) for the \( k \)-th iteration of gradient descent is a metric designed to evaluate the effectiveness of each iteration in reducing the loss function while also considering the stability of the optimization process. The score is given by:

\[
E_k = 100 - 1\min\left(100, \max\left(1, \frac{100 \times P_k}{1 + \log\left(1 + \Delta_k^2\right)}\right)\right)
\]

Let’s break down and explain the components of this formula:

\subsubsection{Logarithmic Term \( \log\left(1 + \Delta_k^2\right) \)}

The term \( \log\left(1 + \Delta_k^2\right) \) serves to dampen the impact of large changes in the loss function.
\begin{itemize}
    \item The logarithm \( \log(x) \) grows slowly as \( x \) increases, so applying it to \( 1 + \Delta_k^2 \) helps moderate large values of \( \Delta_k^2 \). This means that even if \( \Delta_k^2 \) is large, its effect on the efficiency score \( E_k \) will not be excessively amplified.
    \item Adding 1 inside the logarithm ensures that the term \( \log\left(1 + \Delta_k^2\right) \) remains positive, preventing any undefined or negative logarithmic values.
    \item \( \Delta_k^2 \) (the squared difference between successive errors) reflects the stability of the optimization. If the error changes drastically between iterations (i.e., \( \Delta_k \) is large), the logarithmic term will increase, thereby reducing the efficiency score.
\end{itemize}

\subsubsection{The Term \( 1 + \log\left(1 + \Delta_k^2\right) \) in the Denominator}

The purpose of placing \( 1 + \log\left(1 + \Delta_k^2\right) \) in the denominator is to penalize instability in the optimization process.

\begin{itemize}
    \item When the change in the loss function (\( \Delta_k \)) is small, \( \log\left(1 + \Delta_k^2\right) \) will also be small, meaning the efficiency score \( E_k \) will not be heavily penalized.
    \item Conversely, if the loss function change is large, the logarithmic term will increase, leading to a larger denominator and thus a lower efficiency score \( E_k \). This reduction reflects the inefficiency introduced by instability in the optimization process.
\end{itemize}

\subsubsection{Proportional Term \( 100 \times P_k \) in the Numerator}

The term \( 100 \times P_k \) in the numerator represents the proportion of the initial error that has been reduced.

\begin{itemize}
    \item \( P_k \) is the fraction of the initial error that has been reduced by the \( k \)-th iteration, and multiplying it by 100 converts this fraction into a percentage.
    \item This term rewards the optimization process for effectively reducing the error. The larger the error reduction \( P_k \), the higher the numerator, and consequently, the higher the efficiency score \( E_k \).
\end{itemize}

\subsubsection{The Role of \( \min(100, \cdot) \) and \( \max(1, \cdot) \)}

The \( \min(100, \cdot) \) and \( \max(1, \cdot) \) functions ensure that the efficiency score \( E_k \) stays within a reasonable and interpretable range.

\begin{itemize}
    \item \( \max(1, \cdot) \) ensures that the efficiency score does not drop below 1, which prevents the score from becoming too punitive, especially when the change in loss function \( \Delta_k \) is very large.
    \item \( \min(100, \cdot) \) caps the efficiency score at 100, indicating that a score of 100 is the maximum achievable, representing an ideal iteration where error reduction is perfect and stability is maintained.
\end{itemize}

\subsection{Final Function}

The efficiency score \( E_k \) for the \( k \)-th iteration of gradient descent is defined to quantify the effectiveness of each iteration in reducing the loss function while accounting for the stability of the optimization process. The score is given by:

\[
E_k = 100 - \min\left(100, \max\left(1, \frac{100 \times P_k}{1 + \log\left(1 + \Delta_k^2\right)}\right)\right)
\]

Substituting the definitions of \( P_k \) and \( \Delta_k \) into the formula:

\[
E_k = 100 - \min\left(100, \max\left(1, \frac{100 \times \frac{L_{\text{initial}} - L_k}{L_{\text{initial}}}}{1 + \log\left(1 + \left(L_{k-1} - L_k\right)^2\right)}\right)\right)
\]

Next, simplifying the fraction:

\[
E_k = 100 - \min\left(100, \max\left(1, \frac{100 \times (L_{\text{initial}} - L_k)}{L_{\text{initial}} \times \left(1 + \log\left(1 + \left(L_{k-1} - L_k\right)^2\right)\right)}\right)\right)
\]

\section{Experimental Validation}
\subsection{Assumptions and Standards}

The experiments were conducted under the following assumptions:

\begin{itemize}
    \item The loss function being used is ``Mean Squared Error"
    \item The learning rate and other hyperparameters are fixed during each individual experiment and between experiments if relative efficiency is to be taken into account
    \item The datasets used are representative of common machine learning tasks
    \item Standard practices in machine learning, such as using a consistent validation set to monitor performance and ensuring reproducibility through fixed random seeds
\end{itemize}

\subsection{Assumptions and Standards}
The \texttt{generate\_data} function used in this paper produces a synthetic dataset suitable for testing linear regression models and gradient descent optimization techniques.

\subsubsection{Description of the Generated Data}

\begin{itemize}
    \item \textbf{Features} (\( X \)):
    \begin{itemize}
        \item An \( n \times m \) matrix \( X \), where each element is a random value in the range \([0, 2)\).
        \item \( n \) is the number of samples, and \( m \) is the number of features.
        \item The features are generated independently and uniformly at random.
    \end{itemize}

    \item \textbf{Labels} (\( y \)):
    \begin{itemize}
        \item A vector \( y \in \mathbb{R}^n \) where each label is generated based on a linear relationship with the first feature of \( X \), plus some Gaussian noise.
        \item The label for the \( i \)-th sample is computed as:
        \[
        y_i = 4 + 3 \cdot x_{i1} + \epsilon_i
        \]
        where:
        \begin{itemize}
            \item \( x_{i1} \) is the first feature of the \( i \)-th sample.
            \item \( \epsilon_i \sim \mathcal{N}(0, 1) \) is a random noise term drawn from a standard normal distribution.
        \end{itemize}
    \end{itemize}
\end{itemize}

This dataset models a simple linear relationship between the target variable \( y \) and the first feature of \( X \), with added noise to simulate real-world data variability. The remaining features in \( X \) are irrelevant to \( y \), providing a scenario to test feature selection and model robustness.

\subsubsection{Suitability for Gradient Descent Optimization}

Using synthetically generated data like this is highly beneficial for testing gradient descent algorithms due to the following reasons:

\begin{itemize}
    \item \textbf{Known Underlying Model}: Since the true relationship between \( X \) and \( y \) is known, one can easily assess how well the optimization algorithm recovers the underlying parameters (intercept and slope).
    
    \item \textbf{Controlled Noise}: The addition of Gaussian noise allows testing the algorithm's ability to handle data imperfections, which are common in real datasets.
    
    \item \textbf{Feature Irrelevance}: Including multiple features where only one is relevant tests the algorithm's capacity to identify and focus on significant predictors.
    
    \item \textbf{Scalability Testing}: By adjusting \( n \) and \( m \), one can examine the performance and scalability of gradient descent under different dataset sizes.
\end{itemize}

Overall, this synthetic dataset provides a controlled environment to develop, debug, and validate gradient descent optimization techniques before applying them to more complex real-world data.

\subsection{Implementation}

The following function, presented in pseudocode, serves as an example of how to calculate the GDEI in real-time. The objective is to illustrate practical applications of the index and inspire the reader with potential use cases.\newline

\begin{verbatim}
Initialize theta (weights and bias) randomly
Add bias term to input features X_b

SET initial_error = None
Initialize cost_history and efficiency_history

FOR iteration FROM 1 TO n_iterations:
    Predict y_pred using X_b and theta
    Calculate error = y_pred - y
    
    Compute gradients based on error
    Update theta = theta - (learning_rate * gradients)
    
    Calculate current cost (MSE)
    Store current cost in cost_history
    
    IF first iteration:
        SET initial_error = current cost
        CONTINUE
    
    SET prev_cost = cost from previous iteration
    Calculate efficiency using calculate_efficiency(initial_error, cost, prev_cost)
    Store efficiency in efficiency_history
    
    Update learning_rate = learning_rate * decay_rate
\end{verbatim}

\subsection{Plotting the Index}

\begin{figure}[htbp]
\centering
\includegraphics[width=\linewidth]{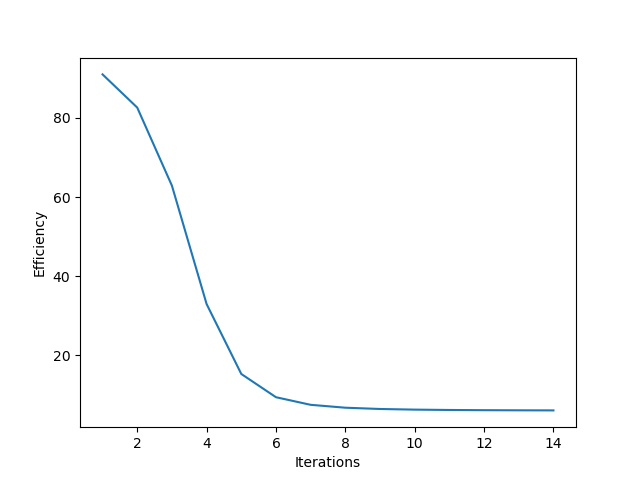}
\caption{Line graph of efficiency with respect to iterations in a sample dataset.}
\label{fig:eplot}
\end{figure}

\hyperref[fig:eplot]{Fig.~\ref*{fig:eplot}} tracks the live efficiency of a randomly generated dataset for linear regression that has an approximate linear curve with noise. The efficiency is between a range of 1 and 100. As show it decreases in each iteration and the rate of change of efficiency (i.e. d\( E_k \)/dk) also retards as the number of iterations progress. 

\section{Conclusion}

This paper introduced a novel efficiency score \( E_k \) for evaluating the effectiveness of each gradient descent iteration. The proposed index captures both the relative reduction in error and the stability of the loss function, offering a more nuanced assessment of gradient descent performance compared to traditional metrics. By providing a finer-grained measure of efficiency, \( E_k \) enables more informed decisions in the selection and adjustment of optimization techniques in machine learning.

The experimental validation demonstrates that \( E_k \) is a robust metric that correlates well with final accuracy and can highlight inefficiencies in the optimization process. Future work will focus on extending the experimental validation of \( E_k \) across a broader range of optimization algorithms, including those with adaptive learning rates and second-order methods. Additionally, the application of \( E_k \) to other optimization problems, such as those involving non-smooth or stochastic loss functions, will be explored. The idea of a "golden number of iterations" or a function that produces the same given a set of parameters can also be explored with this index.

\end{document}